\documentclass[10pt,twocolumn,letterpaper]{article}

\usepackage{iccv}
\usepackage{times}
\usepackage{epsfig}
\usepackage{graphicx}
\usepackage{amsmath}
\usepackage{amssymb}
\usepackage{subfigure}
\usepackage{algorithm}
\usepackage{algorithmic}

\usepackage{amsmath}
\usepackage{graphics}
\usepackage{multicol}

\usepackage[pagebackref=true,breaklinks=true,letterpaper=true,colorlinks,bookmarks=false]{hyperref}

\iccvfinalcopy 

\def\iccvPaperID{2413} 

\begin{document}

\title{Accelerate CNN via Recursive Bayesian Pruning}

\author{Yuefu Zhou, Ya Zhang, Yanfeng Wang \\
Cooperative Medianet Innovation Center, Shanghai Jiao Tong University\\
Shanghai, China\\
{\tt\small \{remicongee, ya\_zhang, wangyanfeng\}@sjtu.edu.cn}
\and
Qi Tian\\
Huawei Noah's Ark Lab\\
China\\
{\tt\small tian.qi1@huawei.com}
}

\maketitle

\begin{abstract}
Channel Pruning, widely used for accelerating Convolutional Neural Networks, is an NP-hard problem due to the inter-layer dependency of channel redundancy. Existing methods generally ignored the above dependency for computation simplicity.
To solve the problem, under the Bayesian framework, we here propose a layer-wise Recursive Bayesian Pruning method (RBP). A new dropout-based measurement of redundancy, which facilitate the computation of posterior assuming inter-layer dependency, is introduced.
Specifically, we model the noise across layers as a Markov chain and target its posterior to reflect the inter-layer dependency.
Considering the closed form solution for posterior is intractable, we derive a sparsity-inducing Dirac-like prior which regularizes the distribution of the designed noise to automatically approximate the posterior. 
Compared with the existing methods, no additional overhead is required when the inter-layer dependency assumed.
The redundant channels can be simply identified by tiny dropout noise and directly pruned layer by layer.
Experiments on popular CNN architectures have shown that the proposed method outperforms several state-of-the-arts.
Particularly, we achieve up to $\bf{5.0\times}$ and $\bf{2.2\times}$ FLOPs reduction with little accuracy loss on the large scale dataset ILSVRC2012 for VGG16 and ResNet50, respectively.
\end{abstract}

\section{Introduction}
Convolutional Neural Networks (CNNs) have recently achieved great success in computer vision and pattern recognition.
However, this success is often accompanied by massive computation which makes the model difficult to deploy on resource-constrained devices.
One popular solution, channel pruning \cite{Alvarez2016LearningTN,Wen2016LearningSS,Louizos2017LearningSN}, lowers computation cost by reducing the number of feature maps.
The key challenge in channel pruning is to identify redundant channels.
Recent Bayesian methods transform variational dropout noise \cite{Kingma2015Variational} as a principled measurement for redundancy via Bayesian inference from sparsity-inducing prior \cite{Louizos2017BayesianCF,Neklyudov2017StructuredBP}.
Redundant channels are considered either being multiplied by a noise of large variance \cite{Louizos2017BayesianCF} or with low Signal-to-Noise Ratio and thus less informative \cite{Neklyudov2017StructuredBP}.
However, these methods assume that the channels in different layers are completely independent and simultaneously infers the redundancy of all layers, which leads to a sub-optimal solution.
In fact, pruning certain channels of any layer is likely to change the distribution of input for the following layer, which may further incite the change of redundancy to fit new input there.
This inter-layer dependency has been considered in heuristics and proved to make pruning more efficient \cite{He2017ChannelPF,Luo2017ThiNetAF}.

In this paper, we attempt to re-investigate the Bayesian pruning framework assuming the inter-layer dependency and propose a layer-wise Recursive Bayesian Pruning method (RBP). 
Similar to existing Bayesian methods~\cite{Neklyudov2017StructuredBP,Louizos2017BayesianCF}, a Gaussian dropout noise, an indicator of channel redundancy, is multiplied on each channel. 
To take the inter-layer dependency into consideration, we model the dropout noise across layers as a Markov chain. The inter-layer dependency is then reflected by the posterior of dropout noise given the dropout noise of the previous layer.
However, the closed form solution for the posterior is intractable. We here derive a sparsity-inducing Dirac-like prior that regularizes the distribution of the dropout noise so as to automatically approximate the posterior. Compared to the existing Bayesian methods, with the Dirac-like prior, RBP requires no additional overhead when assuming the inter-layer dependency.
In addition, the Dirac-like prior is shown to enforce the values of dropout noise to be close to $0$ for redundant channels and close to $1$ for important ones, a desired property of pruning. Thus, we only need to conduct Bayesian inference and prune the channels associated with tiny dropout noise layer by layer.
Additionally, RBP is compatible with reparameterization tricks, which are proved to improve data fitness \cite{Kingma2015Variational}. Hence as a bonus, the performance of CNNs pruned can be recovered fast after a few epochs of finetuning.
In this way, RBP is designed as a completely data-driven approach, achieving a nice balance between data fitness and model acceleration.

We evaluate RBP on popular CNN architectures and benchmark data sets. Our experimental results have shown that RBP outperforms several state-of-the-arts in terms of acceleration. We achieve $\bf{5.0\times}$ and $\bf{2.2\times}$ FLOPs reduction with little accuracy loss on large scale dataset ILSVRC2012 \cite{Deng2009ImageNetAL} for VGG16 \cite{Simonyan2015VeryDC} and ResNet50 \cite{He2016DeepRL}, respectively.


\section{Related work}

Over-parameterization in deep learning often raises huge computation cost, which incites the need for compact neural networks. Pruning is among the most popular solutions in this field and its main idea is removing redundant weights from the original networks.
First introduced in \cite{LeCun1989OptimalBD,Hassibi1992SecondOD}, measurement for redundancy is based on Hessian of the objective function. 
\cite{Han2015LearningBW,Han2016DeepCC} later propose to regard small-magnitude weights as less informative and should be pruned.
However, these methods are unstructured and retain the format of weight matrix, thus the acceleration effect is limited unless Compressed Sparse Column (CSC) adopted. 

Given that, recent trend is pruning whole channels or neurons. \cite{Wen2016LearningSS} proposes group sparsity regularization on weights. \cite{He2017ChannelPF} combines $l_1$-norm regularization and reconstruction error. \cite{Luo2017ThiNetAF} prunes less informative channels layer by layer. \cite{Huang2018DataDrivenSS} extends to select more general structures as residual blocks. Both \cite{He2017ChannelPF,Luo2017ThiNetAF} also consider the influence for redundancy when the input is changed by pruning the previous layer. Particularly, they attempt to suppress this change via minimize a regression loss. By contrast, we propose to infer this change and guide it towards higher sparsity in each layer.

In line with these heuristics, under Bayesian framework, variational dropout \cite{Kingma2015Variational} is adopted to infer the redundancy. \cite{Louizos2017BayesianCF} estimates redundancy from horseshoe prior. \cite{Neklyudov2017StructuredBP} proposes log-normal prior for regularization. Although the proposed method also adopts variational dropout for approximate inference over redundancy, we attempt to tackle the inter-layer dependency and the existing Bayesian methods solely suppose the channels are all independent in networks.

Alternative solutions for compact networks include: 1) Quantization \cite{Rastegari2016XNORNetIC,Chen2015CompressingNN,Wu2016QuantizedCN} reduces bit number of weights stored. 2) Low-rank approximation \cite{Denton2014ExploitingLS,Sindhwani2015StructuredTF,Zhang2016AcceleratingVD} decomposes weight matrix by two stacked smaller ones. 3) Architecture learning \cite{Zoph2017NeuralAS} directly searches compact designs.

\section{Recursive Bayesian Pruning}

\begin{figure}
    \centering
    \subfigure[Process of pruning.]{
    \begin{minipage}[b]{0.46\textwidth}
        \centering
        \includegraphics[width=1\textwidth]{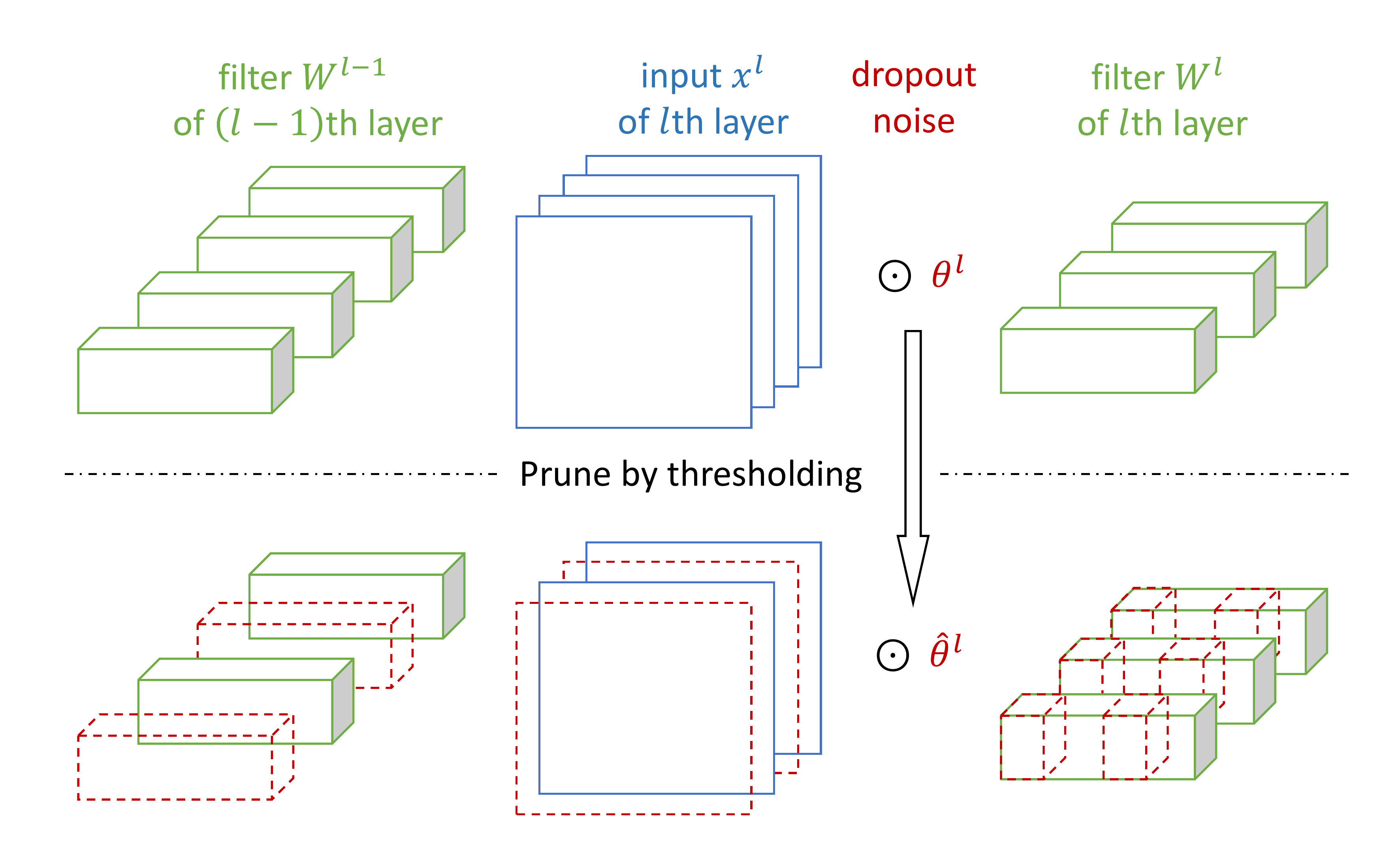} \\
    \end{minipage}
    }
    
    \subfigure[CNN redundancy modeling.]{
    \begin{minipage}[b]{0.46\textwidth}
        \centering
        \includegraphics[width=1\textwidth]{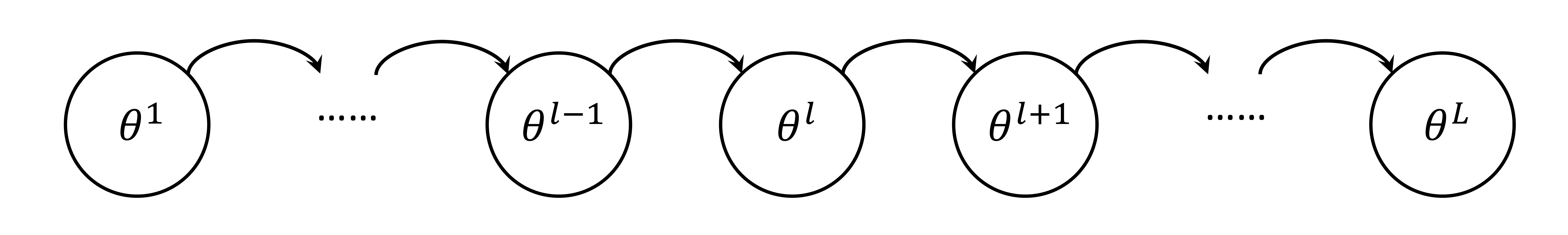} \\
    \end{minipage}
    }
    \caption{Illustration for the proposed method. The pruning is conducted in layer-wise. 
    a) For $l$th layer, we design a vector of dropout noise $\theta^l$ scaled on channels to indicate redundancy.
    The small noise will be assigned as $0$ after estimation done, so that its associated channels and filters are pruned (dotted in red).
    b) While estimating redundancy, given the inter-layer dependency, we model the dropout noise as a Markov chain, thus pruning strategy depends on $p(\theta^l|\theta^{l-1})$.}
    \label{fig: illustration}
\end{figure}

In this section, we provide a comprehensive introduction for the proposed Recursive Bayesian Pruning (RBP) method. Fig. \ref{fig: illustration} provides an illustration of RBP.

We first introduce the notation used in the rest of the paper as follows. $x$ and $y$ represent the data and label sampled from the dataset $D$, respectively.  $x^l$ is the input to $l$th layer and $W^l$ is the filter weight of $l$th layer. $g(.)$ is the activation function. The output of $l$th layer is
\begin{equation}\label{eqn: forward}
    x^{l+1} = W^l * g(x^{l}),
\end{equation}
where $*$ is convolution, and bias is omitted for clarity.

\subsection{Redundancy estimation}

To indicate redundant channels in $x^l$, one intuitive choice is scaling a dropout noise sampled from Bernoulli $\mathcal{B}(1-r)$ \cite{Srivastava2014DropoutAS} with dropout rate $r$ (i.e. the probability of being dropped). 
Given the difficulty of training $r$ under Bayesian framework, we adopt its Gaussian approximation $\mathcal{N}(1-r,r(1-r))$, which is actually the Lyapunov's Central Limit \cite{Wang2013FastDT}.
Let $x^l$ contain $C$ channels, then the Eqn. \ref{eqn: forward} can be rewritten as
\begin{equation}
\begin{split}
    x^{l+1} = W^l * \left(g(x^{l}) \odot \theta^l \right),\;\;\,\\
    \theta^l_c \sim_q \mathcal{N}\left(1-r^l_c, r^l_c(1-r^l_c) \right),
\end{split}
\end{equation}
where $\theta^l=\left(\theta^l_1,...,\theta^l_{C} \right)$, $r^l_c\in[0,1]$ and $\odot$ is element-wise product on channels. In this case, for channels that are probably redundant, i.e. with dropout rate close to $1$, they will be almost pruned since the noise scaling on them is near $0$.
Alternative choices such as log-normal distribution \cite{Neklyudov2017StructuredBP} for the dropout noise is much more complicated than the designed one, but lack an intuitive explanation for redundancy.

To estimate the redundancy with both data fitness and acceleration considered, we maximize the \emph{variational lower bound} $w.r.t.$ $W=\{W^l\}$ and $r^l_c$s:
\begin{equation}\label{eqn: variational lower bound}
    \mathcal{L} = \log\mathbb{P}(y|x,r^l, W) - D_{KL}\left(q(\theta^l)\,||\,p(\theta^l) \right),
\end{equation}
where the first term is log-likelihood, the second term is the Kullback-Leibler (KL) divergence from the estimator $q$ to the sparsity-inducing prior $p$. This training process is equivalent to conduct approximate Bayesian inference on $\theta^l$, and for $W$, we leave it as the optimum for the log-likelihood.

\subsection{Posterior of redundancy}\label{sec: posterior}

Before choosing a sparsity-inducing prior $p$, we return to the core problem: the posterior of redundancy. Recall the observation that pruning channels of one layer may change the input of the following layer, which takes the risk of ruining data fitness. Thus it is preferred to continue pruning and retraining for adaptive weights when knowing how many channels are pruned in the previous layer. In our case, $\theta^l$ indicates redundancy, hence the posterior of redundancy is formed as 
$p(\theta^l|\theta^{l-1})$ for $l$th layer. Directly solving its closed form is difficult, because generally, we can only write the equation below
\begin{equation}
    q(\theta^l) = \int p\left(\theta^l|\theta^{l-1}\right) q(\theta^{l-1})d\theta^{l-1}.
\end{equation}
While seeking for an efficient approximation, we note that once $\theta^{l-1}$ approaches Dirac distribution, the solution is immediate:
\begin{equation}
    \begin{aligned}
        q(\theta^l) \approx& \int p\left(\theta^l|\theta^{l-1}\right) \delta(\theta^{l-1})d\theta^{l-1} \\
                    =&\, p\left(\theta^l|\theta^{l-1}=\mathbb{E}\left[\theta^{l-1}\right]\right).
    \end{aligned}
\end{equation}
This approximation is valid when the Gaussian noise $\theta^{l-1}$ has the dropout rate close to $0$ or $1$. This is intuitively true, because for a highly compact CNN, the channels left are supposed to be important and thus should have tiny probability of being dropped (i.e. $r^{l-1}\approx0$), and for those pruned, once the accuracy is acceptable, there is no reason to keep them (i.e. $r^{l-1}\approx1$). The experiments verify this conjecture, as seen in dropout noise analysis of Section \ref{sec: dropout rate}.

Given that, we simply choose a Dirac-like prior $\mathcal{N}(0,\epsilon^2)$ as the sparsity-inducing prior, where $\epsilon$ is very tiny. Then the KL-divergence in Eqn. \ref{eqn: variational lower bound} can be developed as
\begin{equation}\label{eqn: kl-divergence}
\begin{aligned}
    & D_{KL}\left(q(\theta^l)\,||\,p(\theta^l) \right) \\
     =&\, \sum_{c=1}^C D_{KL}\left(q(\theta^l_c)\,||\,p(\theta^l_c) \right) \\
     =&\, \sum_{c=1}^C -\frac{1}{2}\log\frac{r^l_c(1-r^l_c)}{\epsilon^2} + 
    \frac{1-r^l_c}{2\epsilon^2} - \frac{1}{2}.
\end{aligned}
\end{equation}
Here we adopt mean field theory \cite{Peterson1987AMF} to ease the computation, which supposes the independence among channels within each layer.

Since maximizing the variational lower bound (Eqn. \ref{eqn: variational lower bound}) partially minimizes $D_{KL}$ (Eqn. \ref{eqn: kl-divergence}), the sparsity will be induced by pushing dropout rates to $1$. In fact, let the gradient of $D_{KL}$ $w.r.t.$ $r^l_c$ be zero, i.e. $\partial D_{KL}/\partial r^l_c=0$ , its optimum lies at
\begin{equation}\label{eqn: optimal rate}
    {r^l_c}^* = \frac{1-4\epsilon^2+\sqrt{1+16\epsilon^4}}{2}\approx 1-2\epsilon^2.
\end{equation}

\subsection{Data-driven pruning}

To conduct pruning with data fitness considered, we adopt reparameterization tricks \cite{Kingma2015Variational} by sampling the dropout noise as
\begin{equation}\label{eqn: sample}
    \theta^l_c = 1 - r^l_c + \sqrt{r^l_c(1 - r^l_c)}\cdot\mathcal{N}(0,1),
\end{equation}
which will be scaled on the corresponding channels when forwarding. 
In this way, the dropout rates join the optimization of the log-likelihood (in Eqn. \ref{eqn: variational lower bound}) and can be simply updated via gradient-based strategies. Since the log-likelihood indicates how well the networks fit data, the proposed pruning method is data-driven.

In this paper, we adopt mini-batch update strategy for training each layer. We summarize that on $l$th layer, the objective function to maximize for each batch is
\begin{equation}\label{eqn: objective}
    \mathcal{L} = \mathcal{L}_D - D_{KL}\left(q(\theta^l)\,||\,p(\theta^l) \right),
\end{equation}
where
\begin{equation}
    \mathcal{L}_D = \frac{|D|}{|B|} \sum_{(x,y)\in B} \log\mathbb{P}\left(y|x, W, r^l \right),
\end{equation}
with $B$ a mini-batch. 
At the convergence of this objective, $r^l$ is near $0$ or $1$ and thus $\theta^l$ approximately follow Dirac distribution. 
According to the deduction of section \ref{sec: posterior}, this property will lead to $q(\theta^{l+1})\approx p\left(\theta^{l+1}|\theta^l=\mathbb{E}[\theta^l]\right)$, which incites us to conduct Bayesian inference on $\theta^{l+1}$ with $\theta^{l}$ fixed as its expectation.
Furthermore, given $\mathbb{E}[\theta^l]=1-r^l$ is already near $0$ or $1$, we are free to let large $r^l_c$s be $1$ by thresholding without influencing much the output of this layer. An immediate benefit is that the channels and associated filters of the $l$th and $(l+1)$th layer are directly pruned. To avoid additional cost for storing parameters, we scale the dropout rates on filters
\begin{equation}
\begin{aligned}\label{eqn: threshold and scale}
    r^l_c \leftarrow& 1,\,\text{if}\;\;r^l_c > T \\
    W^{l}_c \leftarrow& W^{l}_c \odot (1 - r^l_c),
\end{aligned}
\end{equation}
where $T$ is threshold value and $W^{l}_c$ is the column for $c$th input channel. Note that $\theta^l$ can be discarded since then.

\subsection{Scale to ResNet}\label{sec: resnet}

\begin{figure}
    \centering
    \includegraphics[width=0.46\textwidth]{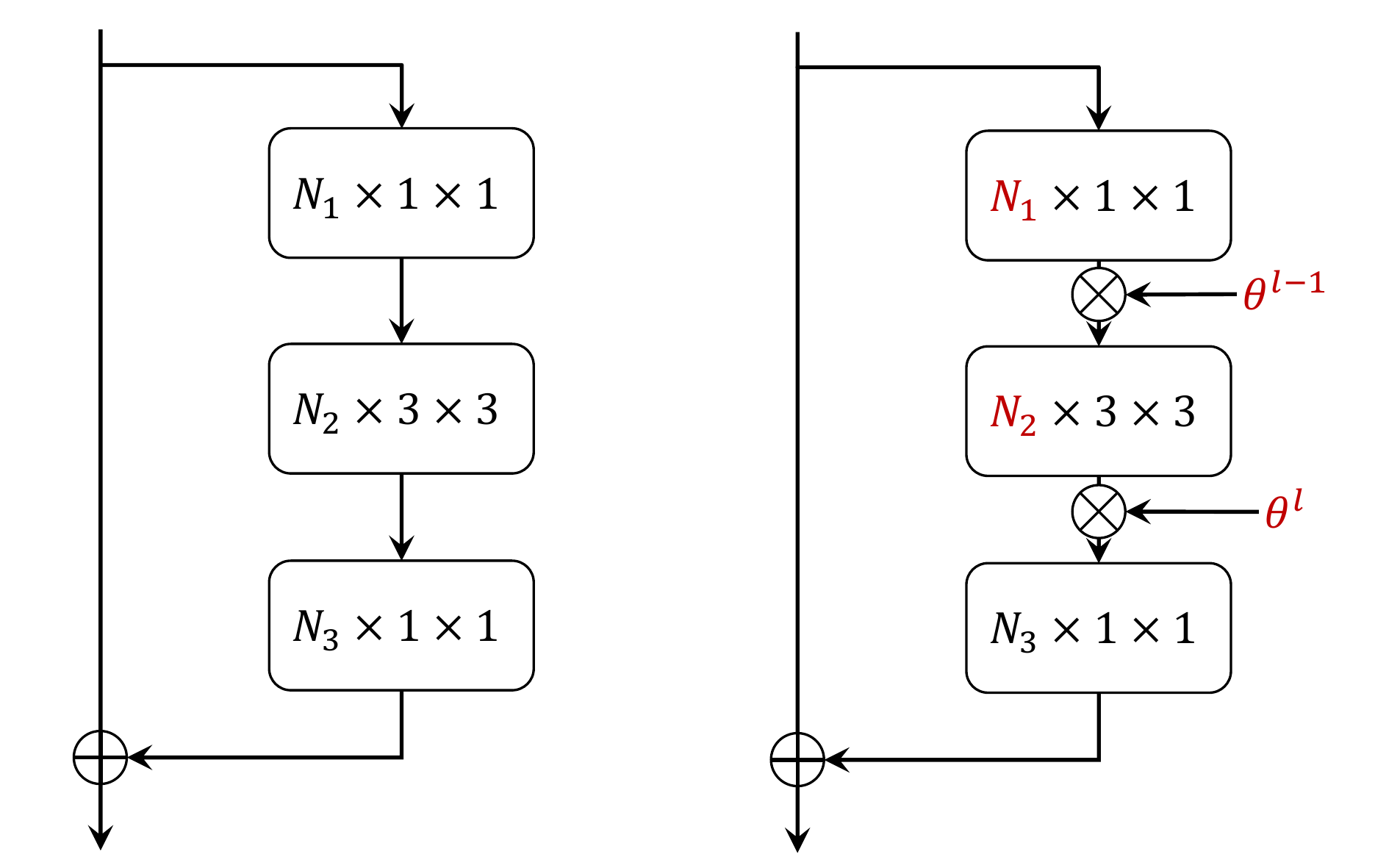}
    \caption{Illustration for pruning residual blocks. The dropout noise is only scaled on the input channels of the last two layers, and thus prunes the filters of the first two layers (in red). $N_1$ and $N_2$ are the number of filters.}
    \label{fig:residual}
\end{figure}

The proposed method can also be applied on residual networks \cite{He2016DeepRL}. As seen in Fig. \ref{fig:residual}, we scale dropout noise on the input of the last two stacked convolutional layers, thus only prune the filters of the first two layers. 
This pruning strategy is also adopted in \cite{Luo2017ThiNetAF,Lin2018AcceleratingCN}, because the output of the last layer is supposed to have the same channel numbers as the input of its residual block so that the sum operation can be valid.

\section{Experiment}

In this section, we validate the effectiveness of the proposed method RBP. The CNN architectures to prune include VGG16 \cite{Simonyan2015VeryDC} and ResNet50 \cite{He2016DeepRL}. We mainly report floating operations (FLOPs) to indicate acceleration effect. Inference-time and storage saving on GPU is measured for practical speed-up results. Compression rate (CR) is also revealed as another criterion for pruning. To have an insight on pruning result, we provide number of channels left.

\begin{algorithm}
\caption{RBP for the whole network.} 
\label{alg: algorithm} 
\begin{algorithmic}[1] 
\REQUIRE 
Dataset $D$, $L$-layer CNN, trigger epoch $E$
\ENSURE 
CNN pruned in channel level 
\WHILE{$l \le L$}
    \STATE $e=0$, $r^l=0.01$, $T=0.5$;
    \FOR{$e$ in range$(E)$}
        \FOR{batch $B$ in $D$}
            \STATE sample $\theta^l$ (Eqn. \ref{eqn: sample});
            \STATE scale $\theta^l$ on input channels (Eqn. \ref{eqn: forward});
            \STATE compute objective $\mathcal{L}$ (Eqn. \ref{eqn: objective});
            \STATE update weight $W$ and $r^l$;
        \ENDFOR
    \ENDFOR
    \STATE $r^l\leftarrow 1$ if $r^l>T$ and scale on $W^l$ (Eqn. \ref{eqn: threshold and scale});
\ENDWHILE
\end{algorithmic} 
\end{algorithm}

\subsection{Implementation}

Given the layer-wise pattern adopted, the condition for moving to the next layer is important. One may choose the moment that the dropout rates are barely updated. In this paper, we observe that the number of epochs required for the convergence is almost the same for all the layers in one architecture. Thus, we simply set one ``trigger epoch" as the number of epochs for training each layer. This value may vary from dataset or architectures, which will be specified later. 
There are two hyper-parameters left to be determined, threshold for pruning $T$ and variance for prior $\epsilon^2$. For the former, since the dropout rates are close to $0$ or $1$, any value in $[0.1,0.9]$ works and does not differ much the results. In this paper, we adopt $T=0.5$. For the later, a dropout rate above $0.95$ almost prunes the corresponding channel, hence it is expected that $1-2\epsilon^2=0.95$ (Eqn. \ref{eqn: optimal rate}) and thus $\epsilon^2=0.025$. A smaller $\epsilon^2$ can be tried for a higher dropout rate for redundant channels. We adopt Adam \cite{Kingma2015AdamAM} as optimizer for RBP and SGD \cite{Sutskever2013OnTI} for finetuning $10$ epochs after pruning. Learning rate is always $1e-4$ during training and degraded by $0.5$ every $3$ epochs during finetuning. For stability of stochastic methods \cite{Snderby2016LadderVA}, we adopt pretrained models.
On CIFAR10, we pretrain the models for $100$ epochs.
On ILSVRC2012, we adopt the pretrained models of ThiNet \cite{Luo2017ThiNetAF}. 
The baseline performance is cited from DDS \cite{Huang2018DataDrivenSS} for fair comparison.
More details can be referred in Algorithm \ref{alg: algorithm}.

\subsection{VGG16 on CIFAR10}

We first prune VGG16 network on CIFAR10 \cite{Krizhevsky2009LearningML}, which contains $50,000$ $32\times32$ images in training set and $10,000$ in test set.
The model performance on CIFAR10 is qualified by the accuracy of classifying $10$-class images.
Considering that VGG16 is originally proposed for large scale dataset, the redundancy is very obvious, especially for the top three fully-connected (fc) layers (in $4096$ dimension). Thus, we cut one fc layer and reduce the dimension of the rest to $512$. To show that RBP can also be applied on fc-layers, we conduct pruning for the whole network, i.e. $13$ convolutional layers and $2$ fc-layers. The trigger epoch for each layer is set as $10$ and the batch size is always $64$.


\begin{table*}
    \begin{center}
        \begin{tabular}{|c|c|c|c|c|c|}
            \hline
            Method & Architecture & CR & FLOPs & Err. \\
            \hline\hline
            Baseline & \textbf{64-64-128-128-256-256-256-512-512-512-512-512-512}-512-512 
            & $1.0\times$ & $1.0\times$ & $8.4$ \\
            SBP \cite{Neklyudov2017StructuredBP} (impl.) & \textbf{47-50-91-115-227-160-50-72-51-12-34-39-20}-20-272 & $17.0\times$ & $3.2\times$ & $9.0$ \\
            BC \cite{Louizos2017BayesianCF} & \textbf{51-62-125-128-228-129-38-13-9-6-5-6-6}-6-20 & $18.6\times$
            & $2.6\times$ & $9.0$ \\
            RBC & \textbf{43-62-120-120-182-113-40-12-20-11-6-9-10}-10-22 & $25.7\times$
            & $3.1\times$ & $9.5$ \\
            IBP & \textbf{45-63-122-123-246-199-82-31-20-17-14-14-31}-21-21 & $13.3\times$ & $2.3\times$ & $8.3$ \\
            \textbf{RBP} & \textbf{50-63-123-108-104-57-23-14-9-8-6-7-11}-11-12 & \textbf{39.1}$\boldsymbol{\times}$ 
            & \textbf{3.5}$\boldsymbol{\times}$ & \textbf{9.0}\\
            \hline
        \end{tabular}
    \end{center}
    \caption{Comparison of pruning VGG16 on CIFAR10. Convolutional layers are in bold. ``impl." denotes our implementation.}
    \label{tab: channel}
\end{table*}

We also duplicate BC \cite{Louizos2017BayesianCF} and SBP \cite{Neklyudov2017StructuredBP} for comparison. These two methods also adopt dropout noise and conduct Bayesian inference for pruning based on different sparsity-inducing prior. BC proposes horeshoe prior in hierarchical form and SBP does log-normal prior.
However, both of them ignore the inter-layer dependency and prune all the channels at the same time.

As seen in Table \ref{tab: channel}, compared with the baseline model, RBP achieves $3.5\times$ FLOPs reduction with only $0.6\%$ error increased. SBP and BC control the error in the same level with RBP, but the acceleration effect is rather modest. In terms of compression rate, RBP exceeds SBP and BC by over $2$ times.
\\[0.5\baselineskip]
{\noindent\large\bf Channel analysis}
\\[0.5\baselineskip]
To have an insight on pruning results, we report the channels left, as seen in Table \ref{tab: channel}. One may wonder that the superior effectiveness of RBP to BC and SBP stems from the layer-wise greedy strategy adopted, hence we implement, for ablation study, 1) IBP, pruning all layers independently at the same time with the proposed objective (Eqn. \ref{eqn: objective}), 2) RBC, applying BC layer by layer.
Comparing BC and RBP, we note that both BC and RBP prune $\texttt{conv5}$ (last three convolutional layers) to very few channels ($\sim10$). However, RBP is able to prune more $\texttt{conv1}$-$\texttt{conv4}$, where lies about $90\%$ FLOPs. 

By adopting layer-wise strategy, RBC improves dramatically the compression rate and FLOPs reduction, but with more error increased. We suppose that in the theory of BC, the redundancy estimator and prior are neither designed for inter-layer dependency. Although pruning layer by layer, the distribution of BC's dropout noise can not fit the posterior of redundancy and thus may prune more in each layer but misunderstand the distribution of input.

Another interesting observation is that applying RBP for all the layers at the same time lets the pruning result approach BC. For instance, there is less difference between IBP and BC in $\texttt{conv2}$, i.e. $246\,v.s.\,228$ and $199\,v.s.\,129$. We ascribe the performance loss to the absence of layer-wise strategy. Pruning layer by layer in data-driven way can ``inform" the following layers that data fitness can be retained with less filters. In this case, both BC and IBP keep most channels in $\texttt{conv1}$ and $\texttt{conv2}$ but still fail to reduce redundancy in the following layers. By contrast, RBP prunes to $104/256$ and $57/256$ channels in $\texttt{conv3}$.

\subsection{VGG16 on ILSVRC}

We now evaluate the performance of RBP for VGG16 on ILSVRC2012 \cite{Deng2009ImageNetAL}. ILSVRC2012 is a large-scale image classification dataset, which contains $1,000$ classes, more than $1.2$ million images in training set and $5,0000$ in validation set. As input of VGG16, we sample $128$ images as a batch and adopt data augmentation for each one when training: 1) resize to $256\times256$ and crop randomly a $224\times224$ part, 2) adopt random horizontal flip, 3) normalize with mean value and standard deviation pre-defined. During test, we almost apply the same data augmentation, except that a $224\times224$ part is extracted in the center.
For VGG16 in this section, we return to the original architecture, i.e. $13$ convolutional layers and $3$ $4096$-d fc layers.

In terms of pruning strategy, we do not prune the whole network this time. Instead, the first $10$ convolutional layers ($\texttt{conv1}$-$\texttt{conv4}$) are to be pruned. This strategy is commonly adopted for VGG16 on ILSVRC, because as mentioned before, more than $90\%$ FLOPs is distributed on these layers. Given that, we apply RBP on the first $10$ layers during $30$ epochs and the trigger epoch is thus $3$. And since we do not prune the fc layers, which contains most parameters, we focus on the FLOPs reduction.

We compare the results with DDS \cite{Huang2018DataDrivenSS}, ThiNet \cite{Luo2017ThiNetAF} and CP \cite{He2017ChannelPF}. Similar with us, DDS also adopts a scale value to indicate redundancy, while the regularization is heuristic. ThiNet and CP also consider the influence for redundancy when the input is changed by pruning the previous layer. Particularly, they attempt to suppress this change via minimize a regression loss. By contrast, we propose to infer this change and guide it towards higher sparsity in each layer.
All of these three methods prune the first $10$ layers as ours, except that CP set manually the number of remaining channels.

As shown in Table \ref{tab: vgg16}, RBP reduces FLOPs by $5\times$ and still achieves competitive accuracy. Compared with DDS, RBP shows superior speed-up ratio with $0.1\%$ top-1 error increased and $0.4\%$ top-5 accuracy improved. Compared with CP, RBP does not only achieve lower FLOPs but also provides better classification accuracy.
CP outperforms ThiNet on FLOPs reduction but with $2\%$ top-1 error increased. The gap mainly stems from the factitious pruning settings of CP, while ThiNet simply prunes half channels for each layer. We suppose that both these strategies are not effective enough. For CP, manually setting the remaining channel ratios introduces hyper-parameters and may need additional cost for tuning. Although this helps CP progressively prunes networks, the accuracy loss is also obvious because the hyper-parameters' setting is not data-driven. For ThiNet, pruning uniformly all the layers ignores the possibility that redundancy varies from depth. In fact, it has been widely known that deeper layers extract higher level semantic information, thus different functionality may require different numbers of filters for data fitness. 

\begin{table}
    \begin{center}
        \begin{tabular}{|c|c|c|c|}
        \hline
         Method & FLOPs & Top-1 Err. & Top-5 Err. \\
         \hline\hline
         Baseline & $1.0\times$ & $27.5$ & $9.2$ \\ 
         DDS \cite{Huang2018DataDrivenSS} & $4.0\times$ & $31.5$ & $11.8$  \\
         ThiNet \cite{Luo2017ThiNetAF} & $3.2\times$ & $30.2$ & $10.5$ \\
         CP \cite{He2017ChannelPF} & $4.4\times$ & $32.2$ & $11.9$ \\
         {\bf RBP} & $\bf{5.0\times}$ & $\bf{31.6}$ & $\bf{11.4}$ \\
         \hline
    \end{tabular}
    \end{center}
    \caption{Comparison results of VGG16 on ILSVRC2012. Top-$k$ Err. denotes the classification error for the first $k$ predictions.}
    \label{tab: vgg16}
\end{table}

We also report the number of remaining channels for $\texttt{conv1}$-$\texttt{conv4}$ in Table \ref{tab: vgg16 channel}. Totally, we prune around two-thirds of channels. In channel level, one interesting observation is that the last layer of every $\texttt{conv}$ block is pruned to around half channels, while the rest are reduced to around one quarter. Why is it harder to prune the former? We attribute it to the sensitivity raised by resolution reduction. In fact, the last layer of each block is stacked by a pooling layer to reduce size of feature maps. For instance, the feature maps generated by $\texttt{conv2\_2}$ are sampled from $112\times112$ to $56\times56$.
Thus, to maintain enough information, more channels may be required by the following block.
This gives us a clue that pruning all the layers with the same remaining ratio, such as in ThiNet, is unwise, which may take the risk of pruning too much for sensitive layers or leaving much redundancy for others.
Additionally, the FLOPs reduction accumulated in Table \ref{tab: vgg16 channel} may provide useful suggestions for pruning strategy. Note that FLOPs is reduced most in $\texttt{conv4}$ block. Especially on $\texttt{conv4\_2}$ layer, the speed-up ratio grows from $2.9\times$ to $4.3\times$. We suppose that there exits most redundancy in this block. 

\begin{table}
    \begin{center}
        \begin{tabular}{|c|c|c|c|}
        \hline
         Layer & \#Remained/\#Original &Percent & FLOPs \\
         \hline\hline
         \multicolumn{4}{|c|}{$224\times224$} \\
         \hline
         conv1\_1 & 16/64 & $25\%$ & $1.1\times$\\
         conv1\_2 & 39/64 & $60\%$ & $1.2\times$\\ \hline
         \multicolumn{4}{|c|}{$112\times112$} \\
         \hline
         conv2\_1 & 45/128 & $35\%$ & $1.3\times$\\
         conv2\_2 & 81/128  & $63\%$ & $1.4\times$\\ \hline
         \multicolumn{4}{|c|}{$56\times56$} \\
         \hline
         conv3\_1 & 65/256  & $25\%$ & $1.6\times$\\
         conv3\_2 & 68/256  & $26\%$ & $2.0\times$\\
         conv3\_3 & 116/256  & $45\%$ & $2.2\times$\\ \hline
         \multicolumn{4}{|c|}{$28\times28$} \\
         \hline
         conv4\_1 & 132/512  & $26\%$ & $2.9\times$\\
         conv4\_2 & 135/512  & $26\%$ & $4.3\times$\\
         conv4\_3 & 257/512  & $50\%$ & $5.0\times$\\
         \hline\hline
         Total & 954/2688  & $35\%$ & $5.0\times$ \\
         \hline
    \end{tabular}
    \end{center}
    \caption{Remaining channels of VGG16 on ILSVRC2012. FLOPs reduction is reported in form of accumulation. Resolution of input channels is over each $\texttt{conv}$ block.}
    \label{tab: vgg16 channel}
\end{table}

\subsection{ResNet50 on ILSVRC}\label{sec: resnet imagnet}

We now accelerate ResNet50 on ILSVRC2012. ResNet50 is a very deep CNN in the residual network family. It contains $16$ residual blocks \cite{He2016DeepRL}, where around $50$ convolutional layers are stacked. Although the depth of ResNet50 is greater than VGG16, many filters of the former are of size $1\times1$ and hence already saves much FLOPs, i.e. $4.1$ billion $v.s.$ $31.0$ billion. However, reported in PyTorch model zoo \cite{paszke2017automatic}, ResNet50 outperforms VGG16 by around $3\%$ top-1 accuracy on ILSVRC2012. Given that, we suppose that ResNet50 is already a much more compact architecture than VGG16 and pruning should be more cautious.

In this section, we always follow the strategy proposed in Section \ref{sec: resnet}, i.e. only prune the filters of the first two convolutional layers of each residual block. Furthermore, considering the following factors, we improve RBP to be more adaptive to residual network family:
\begin{itemize}
    \item[1)] Although we choose to only prune the filters of the first two convolutional layers of each residual block, there still exists $32$ layers, which will be exhausting if the trigger epoch is large. Given that, we assume that the dependency across blocks is relatively weak and can be ignored. This is intuitively reasonable, because between two adjacent blocks, the layers to be pruned are separated by another convolutional layer and divided into two groups. Therefore, we are free to prune the first layers of all the blocks at the same time, and then move on all the second layers.
    \item[2)] It has been found that the residual networks are very sensitive at the blocks with down-sampling layers and not robust to pruning \cite{Li2017PruningFF}. In ResNet50, there are $4$ residual blocks containing down-sampling layers. We propose to omit these blocks for better data fitness.
\end{itemize}
We name the variant RBP combined with the above two points ResNet-adaptive RBP (RRBP). The trigger epoch is respectively $3$ and $7$ for RBP and RRBP. Both sample $256$ images as a batch with the same data augmentation as for VGG16.

The pruning results are shown in Table \ref{tab: resnet50}. For ThiNet, we cite ThiNet-50 which prunes $50\%$ channels. And for DDS, we cite DDS(32) and DDS(26), where DDS respectively prunes ResNet50 to $32$ and $26$ residual blocks. We also report the performance of RBP, which simply conducts layer-wise pruning on ResNet50. It can be found that both RBP and RRBP achieve FLOPs reduction over $2\times$, while DDS and CP are rather conservative. 
In terms of classification accuracy, DDS(32) provides the lowest top-1 and top-5 error, however, the speed-up ratio is also the lowest. By contrast, DDS(26) prunes ResNet50 more progressively and outperforms DDS(32). Even so, RBP and RRBP show significant superiority to DDS(26). In fact, the former reduce almost $1\times$ more FLOPs but keep competitive classification accuracy, i.e. RBP holds only $0.7\%$ more top-1 error and RRBP even provides $1.2\%$ less. Given this point, we conclude that RBP and RRBP are more effective on residual networks than DDS. 
Compared with ThiNet-50, RBP and RRBP achieves the same level of FLOPs reduction with competitive classification accuracy. Particularly, RRBP is even $2.0\%$ and $1.0\%$ better on Top-1 and Top-5 accuracy, respectively.
Between RBP and RRBP, the later shows almost the same acceleration effect but with higher classification accuracy, which validates our idea that RRBP is more adaptive to residual networks.

\begin{table}
    \begin{center}
        \begin{tabular}{|c|c|c|c|}
        \hline
         Method & FLOPs & Top-1 Err. & Top-5 Err. \\
         \hline\hline
         Baseline & $1.0\times$ & $23.9$ & $7.1$ \\ 
         DDS(32) \cite{Huang2018DataDrivenSS} & $1.4\times$ & $25.8$ & $8.1$  \\
         DDS(26) \cite{Huang2018DataDrivenSS} & $1.7\times$ & $28.2$ & $9.2$ \\
         CP \cite{He2017ChannelPF} & $1.5\times$ & $27.7$ & $9.2$ \\
         ThiNet-50 \cite{Luo2017ThiNetAF} & $2.3\times$ & $29.0$ & $10.0$ \\
         {\bf RBP} & $\bf{2.3\times}$ & $\bf{28.9}$ & $\bf{10.0}$ \\
         {\bf RRBP} & $\bf{2.2\times}$ & $\bf{27.0}$ & $\bf{9.0}$ \\
         \hline
    \end{tabular}
    \end{center}
    \caption{Comparison results of ResNet50 on ILSVRC2012. Top-$k$ Err. denotes the classification error for the first $k$ predictions.}
    \label{tab: resnet50}
\end{table}

\begin{table}
    \begin{center}
        \begin{tabular}{|c|c|c|c|}
        \hline
         Stage & \#Remained/\#Original &Percent & FLOPs \\
         \hline\hline
         res2 & 20/256 & $9\%$ & $1.2\times$\\ \hline
         res3 & 67/1024 & $7\%$ & $1.6\times$\\ \hline
         res4 & 2408/3072  & $78\%$ & $1.8\times$\\ \hline
         res5 & 1105/3072  & $36\%$ & $2.3\times$\\
         \hline\hline
         Total & 3600/7424  & $48\%$ & $2.3\times$ \\
         \hline
    \end{tabular}
    \end{center}
    \caption{{\bf RBP} result. Remaining channels of ResNet50 on ILSVRC2012. FLOPs reduction is reported in form of accumulation. The \#Remained and \#Original only count the channels in the first two convolutional layers of each residual block. Stage groups residual blocks between two down-sampling layers.}
    \label{tab: rbp channel}
\end{table}

Table \ref{tab: rbp channel} shows the remaining channels in each block when applying RBP. Totally, we prune more than half of channels. In stage level, over $90\%$ channels are pruned in $\texttt{res1}$ and $\texttt{res2}$, yielding most FLOPs reduction contribution. With a close look at $\texttt{res2\_2}$, we find that only $1$ filter is remained in the first two convolutional layers, which almost removes this residual block. Note that DDS is proposed to prune a more general structure rather than channels, such as residual blocks. In this case, RBP simulates block selection by pruning most channels there, which shows a similar generality with DDS. However, most channels of $\texttt{res4}$ are kept, while DDS(32) prunes two residual blocks. This is mainly because simply adopting layer-wise strategy on ResNet50 may over-prune the first several residual blocks and thus requires the more filters in the following layers to ensure data fitness. Furthermore, RBP also progressively prunes the sensitive blocks with down-sampling layers, which explains why the classification error is higher.

\begin{figure}
    \centering
    \includegraphics[width=0.48\textwidth]{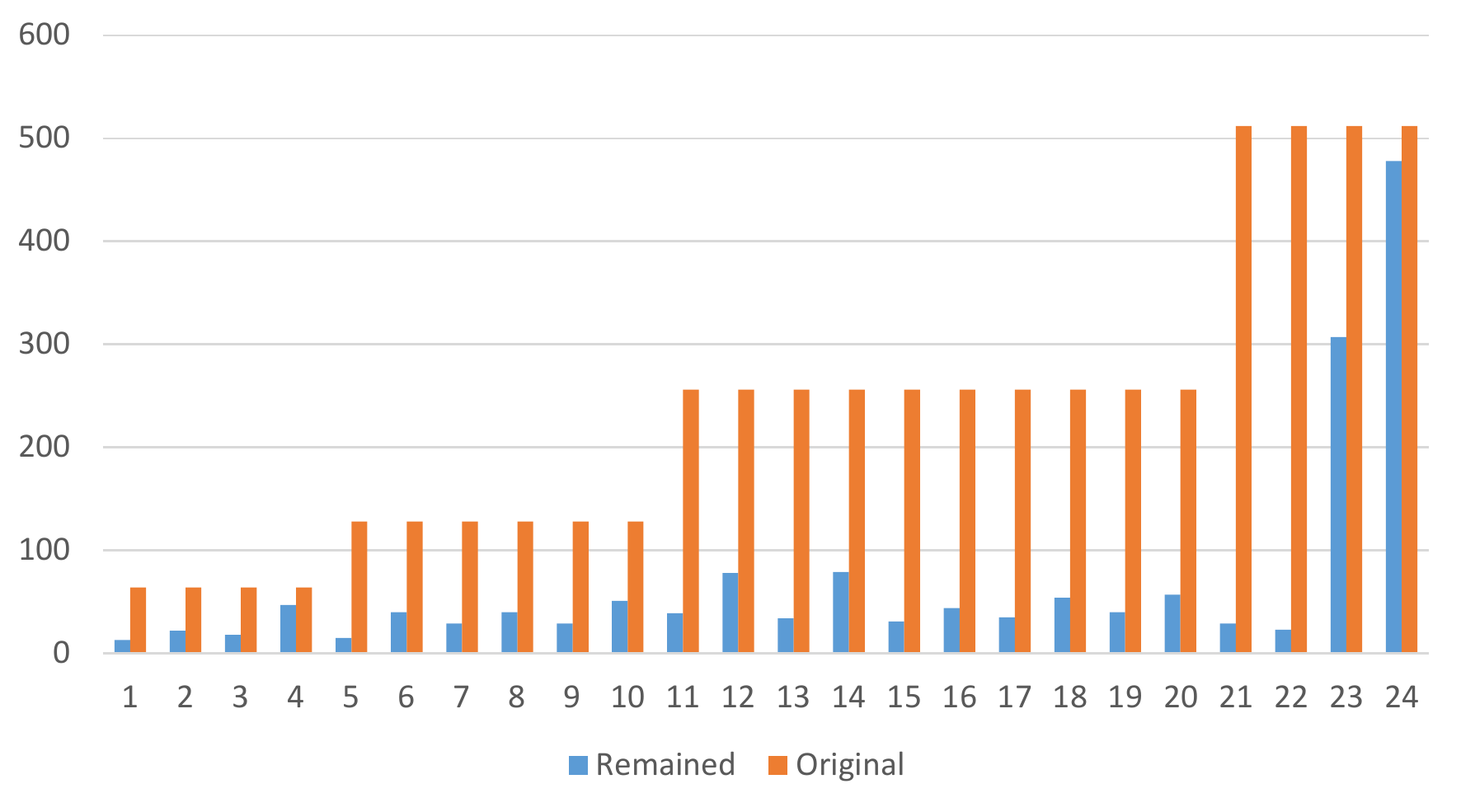}
    \caption{{\bf RRBP} result. Columns for comparison between the number of remaining channels and original ones in each layer.}
    \label{fig: rrbp channel}
\end{figure}

Fig. \ref{fig: rrbp channel} shows the channels remained after pruning by RRBP. It can be easily found that compared with RBP, RRBP prunes channels more uniformly and almost reduces the redundancy of each layer to a very low level. The only exceptions are the convolutional layer in the last residual block ($23$ and $24$ in Fig. \ref{fig: rrbp channel}). We suppose that although this block contains no down-sampling layers, it is stacked by a pooling layer that reduces the resolution from $7\times7$ to $1\times1$. Despite the convenience for the following fc layer, it makes the last residual block more sensitive for channel pruning. This observation is consistent with the result of VGG16, where the layers before pooling retains more channels. For totality, RRBP removes $4,000$ channels, which is even better than RBP ($3824$ channels removed). Note that the pruning field of RRBP is smaller than RBP, because $4$ residual blocks with down-sampling layers are ignored, however, it still shows superior performance for identifying redundant channels. The robustness of RRBP to residual networks is thus an immediate conclusion.

\subsection{Detail analysis}\label{sec: dropout rate}

{\noindent\large\bf Practical acceleration}

\begin{figure}
    \centering
    \subfigure[Dropout rates of VGG16.]{
    \begin{minipage}[b]{0.46\textwidth}
        \centering
        \includegraphics[width=1\textwidth]{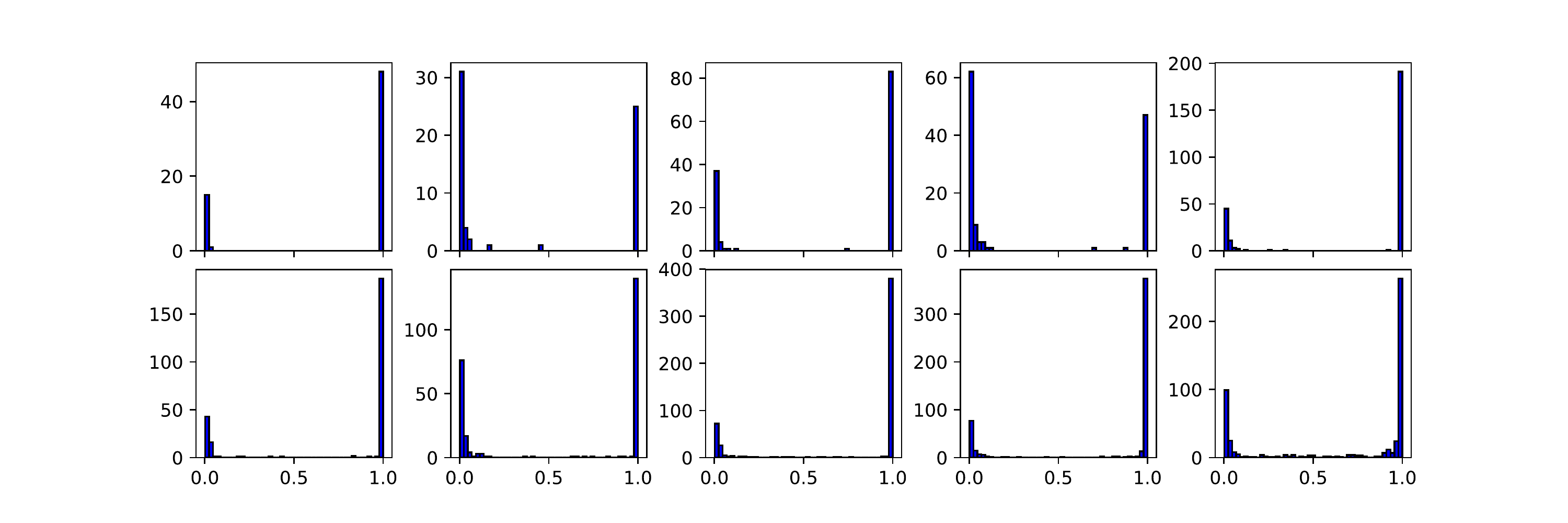} \\
    \end{minipage}
    }
    
    \subfigure[Dropout rates of ResNet50.]{
    \begin{minipage}[b]{0.47\textwidth}
        \centering
        \includegraphics[width=1\textwidth]{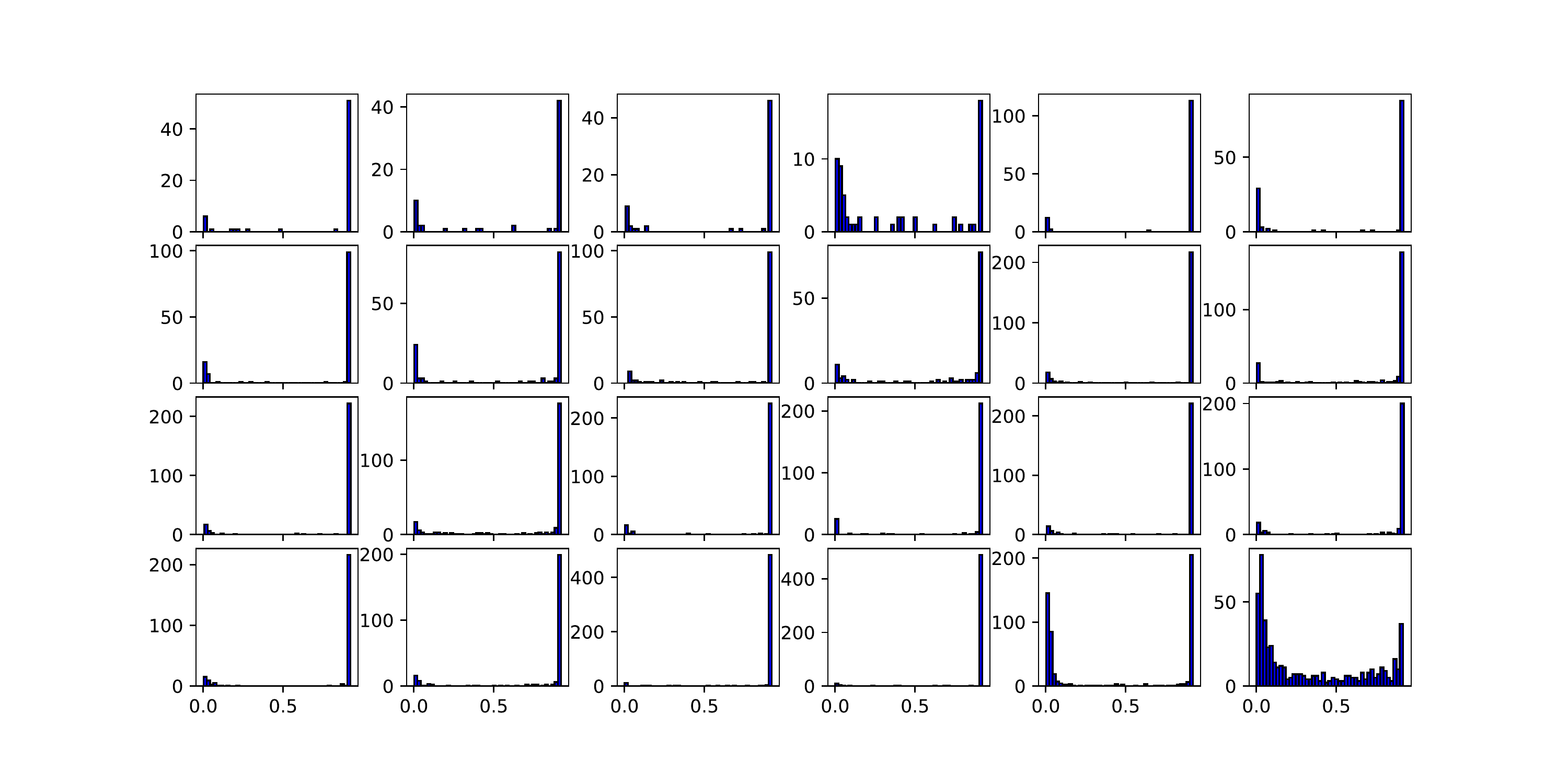} \\
    \end{minipage}
    }
    \caption{Histogram of dropout rates in each layer of VGG16 and ResNet50 (RRBP).}
    \label{fig: droppout rate}
\end{figure}

\begin{table}
    \begin{center}
        \begin{tabular}{|c|c|c|c|}
        \hline
         Model & Time & Storage & Top-5 Err. \\
         \hline\hline
         VGG-RBP & $2.6\times$ & $3.6\times$  & $+2.2$ \\
         ResNet50-RBP & $1.4\times$ & $1.5\times$ & $+2.9$ \\
         ResNet50-RRBP & $1.3\times$ & $1.4\times$ & $+1.8$ \\
         \hline
    \end{tabular}
    \end{center}
    \caption{GPU acceleration for VGG and ResNet50 on ILSVRC.}
    \label{tab: gpu}
\end{table}

For ILSVRC2012, We also evaluate the acceleration performance on GPU (GeForce GTX 1080 Ti). All the models are run under Caffe \cite{Jia2014CaffeCA} with CUDA8 \cite{Nickolls2008ScalablePP} and cuDNN5 \cite{Chetlur2014cuDNNEP}. The inference time is averaged from $50$ runs of batch size $32$.
As shown in Table \ref{tab: gpu}, the proposed method achieves promising acceleration and lower storage with little accuracy drop on VGG16 and ResNet50, respectively.
\\[0.5\baselineskip]
{\noindent\large\bf Fast recover}

\begin{figure}
    \centering
    \includegraphics[width=0.48\textwidth]{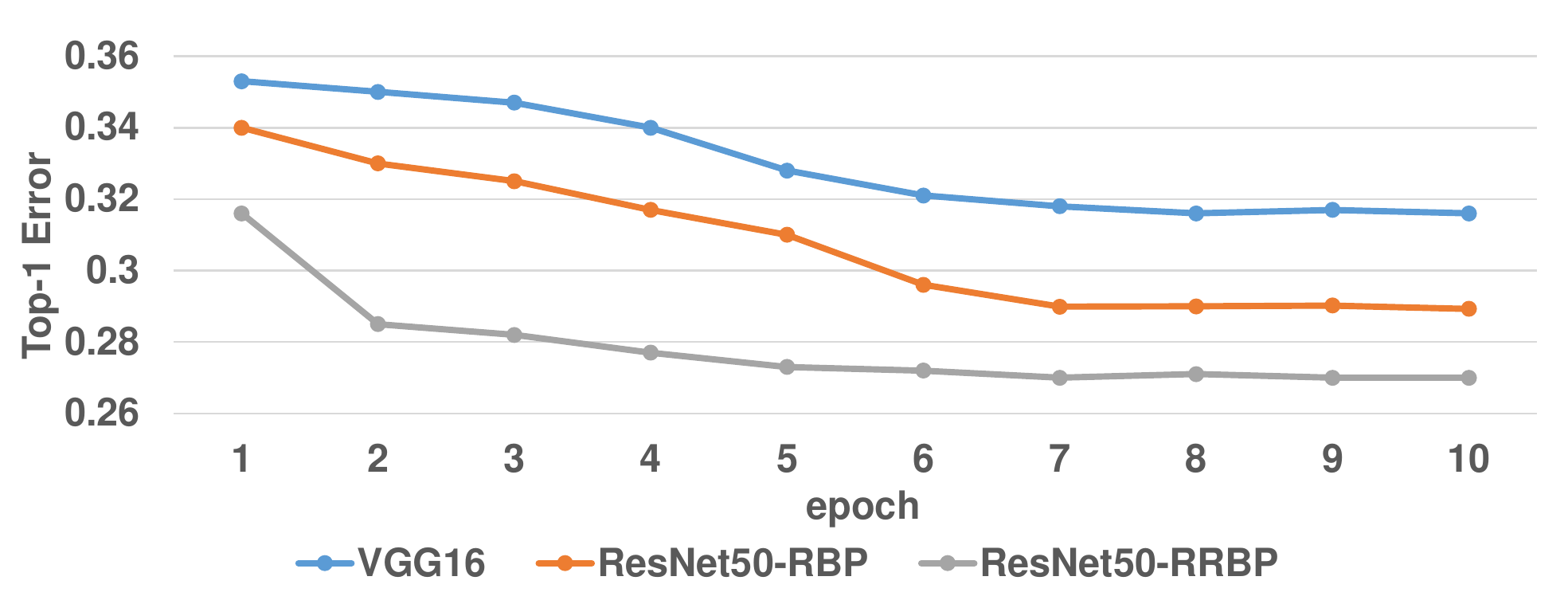}
    \caption{Top-1 error change curve during finetuning on ILSVRC.}
    \label{fig: fintune}
\end{figure}

As the proposed method is claimed to be data-driven, we show that the data fitness is retained much after pruning and can be recovered fast in a few epochs of finetuning. The results are shown in Fig. \ref{fig: fintune}. The convergence occurs at the $7$th epoch. Particularly, ResNet50 pruned by RRBP is recovered the fastest. This is because RRBP does not prune the sensitive residual blocks and thus retain more powerful data fitness.
\\[0.5\baselineskip]
{\noindent\large\bf On dropout noise}

Remind that in Section \ref{sec: posterior}, the dropout rates $r^l$ are supposed to be near $0$ or $1$ after optimization of Eqn. \ref{eqn: objective} done.
This hypothesis is the precondition for the designed dropout noise to approach Dirac distribution. In this section, we provide experimental proofs that this hypothesis is generally valid. As shown in Fig. \ref{fig: droppout rate}, almost all of the dropout rates are distributed near $0$ or $1$. Note that in the last block of ResNet50, some dropout rates does not approach $0$ or $1$, which is consistent with the proposition that this layer is sensitive to pruning (Section \ref{sec: resnet imagnet}).

\section{Conclusion}

In this paper, we extend the existing Bayesian pruning methods by embedding inter-layer dependency.
By proposing RBP, the redundant channels are identified efficiently and directly pruned layer by layer.
Given the data-driven pattern adopted, a nice balance between data fitness and model acceleration is found.
The experiments on popular CNN architectures validate the effectiveness of the proposed method, also showing superior performance to the state-of-the-arts.

{\small
\bibliographystyle{ieee}
\bibliography{egbib}
}

\end{document}